\PassOptionsToPackage{table}{xcolor}
\documentclass[sigconf,review = false]{acmart}
\usepackage[table]{xcolor}
\usepackage{subfig}

\usepackage{amsfonts,amssymb} 
\usepackage{amsmath}
\usepackage{amsthm}
\usepackage{bm}
\usepackage{multirow}
\usepackage{makecell}
\usepackage{arydshln}
\usepackage{booktabs}
\usepackage{multicol}

\AtBeginDocument{%
	}

\setcopyright{acmcopyright}
\copyrightyear{2023}
\acmYear{2023}
\acmDOI{XXXXXXX.XXXXXXX}

\acmConference[MM '23]{Proceedings of the 31th ACM International Conference on Multimedia}{October 28--Novmber 03,
	2023}{Ottawa, Canada}
\acmBooktitle{Proceedings of the 31th ACM International Conference on Multimedia (MM '23), October 28--Novmber 03, 2023, Ottawa, Canada}
\acmPrice{15.00}
\acmISBN{978-1-4503-XXXX-X/18/06}

\begin{document}
	
	\title{Visual Commonsense based Heterogeneous Graph Contrastive Learning}
	
	\author{Zongzhao Li}
	\email{lizongzhao2020@ia.ac.cn}
	\affiliation{%
		\institution{CBSR\&NLPR, Institute of Automation, Chinese Academy of Sciences}
		\institution{School of Artificial Intelligence, University of Chinese Academy of Sciences}
		\city{Beijing}
		\country{China}
	}
	
	\author{Xiangyu Zhu}
	\email{xiangyu.zhu@nlpr.ia.ac.cn}
	\affiliation{%
		\institution{CBSR\&NLPR, Institute of Automation, Chinese Academy of Sciences}
		\institution{School of Artificial Intelligence, University of Chinese Academy of Sciences}
		\city{Beijing}
		\country{China}
	}
	
	\author{Xi Zhang}
	\email{zhangxi2019@ia.ac.cn}
	\affiliation{%
		\institution{NLPR, Institute of Automation, Chinese Academy of Sciences}
		\institution{School of Artificial Intelligence, University of Chinese Academy of Sciences}
		\city{Beijing}
		\country{China}
	}
	
	\author{Zhaoxiang Zhang}
	\email{zhaoxiang.zhang@ia.ac.cn}
	\affiliation{%
		\institution{CBSR\&NLPR, Institute of Automation, Chinese Academy of Sciences; School of Artificial Intelligence, University of Chinese Academy of Sciences; Centre for Artificial Intelligence and Robotics, Hong Kong Institute of Science \& Innovation, Chinese Academy of Sciences}
		\city{Beijing}
		\country{China}
	}
	
	\author{Zhen Lei}
	\email{zlei@nlpr.ia.ac.cn}
	\affiliation{%
		\institution{CBSR\&NLPR, Institute of Automation, Chinese Academy of Sciences; School of Artificial Intelligence, University of Chinese Academy of Sciences; Centre for Artificial Intelligence and Robotics, Hong Kong Institute of Science \& Innovation, Chinese Academy of Sciences}
		\city{Beijing}
		\country{China}
	}
	
	\renewcommand{\shortauthors}{Li et al.}
	\settopmatter{printacmref=false} 
	\begin{abstract}
		How to select relevant key objects and reason about the complex relationships cross vision and linguistic domain are two key issues in many multi-modality applications such as visual question answering (VQA). In this work, we incorporate the visual commonsense information and propose a heterogeneous graph contrastive learning method to better finish the visual reasoning task. Our method is designed as a plug-and-play way, so that it can be quickly and easily combined with a wide range of representative methods. Specifically, our model contains two key components: the Commonsense-based Contrastive Learning and the Graph Relation Network. Using contrastive learning, we guide the model concentrate more on discriminative objects and relevant visual commonsense attributes. Besides, thanks to the introduction of the Graph Relation Network, the model reasons about the correlations between homogeneous edges and the similarities between heterogeneous edges, which makes information transmission more effective. Extensive experiments on four benchmarks show that our method greatly improves seven representative VQA models, demonstrating its effectiveness and generalizability.
	\end{abstract}

	
	
	\maketitle
	
	\section{Introduction}
	Recent years have witnessed tremendous research works developed in the vision-language tasks \cite{li2021exploring,wang2021t2vlad,lei2021less,yang2021deconfounded,su2018learning}. Multiple-choice visual question answering (VQA) \cite{jang2017tgif,xiao2021next} is the typical one of them, which demands the model to select the correct one among several answer candidates given a question clue and visual inputs (an image or a video). This task requires the model to reason about the relations between various visual and textual entities on the cognitive level of vision and linguistic domains. The major challenge of the problem is to build a relatively accurate information flow between low-level visual features (image side) and high-level semantic features (text side). This foundation allows for the ongoing discovery of key entities as well as the iterative updating of relations between entities.
	
\begin{figure*}[t]
		\begin{center}
			\includegraphics[scale=0.10]{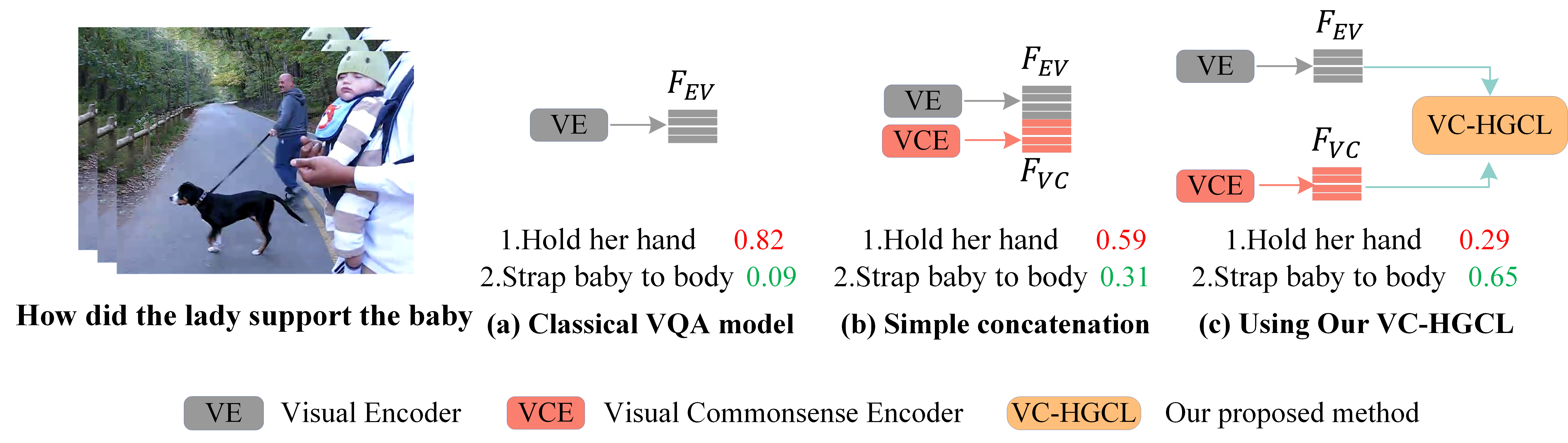}
		\end{center}
		\caption{Experimental analysis of the visual commonsense on NExT-QA \cite{xiao2021next}. As shown in (a), because the visual commonsense is not taken into account, the classical VQA model tends to select "Hold her hand", which is the most prevalent correlation between \texttt{baby} and \texttt{adult's hands}. After accounting for the visual commonsense and concatenating it with the original visual feature (b), the model's prediction probability for option "\texttt{Hold her hand}" has reduced due to the introduction of various sorts of correlation information. When effectively utilizing the visual commonsense (c), our method mines semantic information pertinent to the current circumstance and selects the appropriate choice "\texttt{Strap baby to the body}". }
		\label{figure1}
	\end{figure*}

	Many efforts have been done to model the interaction between video (image) and text in earlier works. However, due to the lack of attention to visual commonsense information, they cannot accurately discover significant objects and create rational relationships. How to utilize visual commonsense information and make reasonable use of it is important in this field. The visual commonsense referenced in this paper is consistent with the definition in \cite{wang2020visual}. Visual commonsense is essentially a visual feature extracted based on causal relation that contains more semantic information. For a visual object, its visual commonsense feature considers its real causal relation with the surroundings in various contexts. We utilize Figure \ref{figure1} as an example to provide a more thorough explanation of the visual commonsense. In real-world scenarios, the baby will typically be surrounded by hands, signifying that \texttt{the adult is holding the baby}; however occasionally, the baby will be surrounded by straps, signifying that \texttt{the adult is strapping baby to the body}. Since the dataset for the model training is drawn from real-world observations, there will be many instances in which \texttt{baby} and \texttt{adult's hands} cohabit in the dataset and few instances in which \texttt{baby} and \texttt{strap} coexist. As a result, during the training phase, the model gives higher weight to the semantic information of \texttt{strap} rather than the semantic information of \texttt{adult's hands} in the \texttt{baby}'s visual feature. Hence, when faced with the situation in Figure \ref{figure1} and the question "How did the lady support the baby?", the classical VQA model in Figure \ref{figure1} (a) disregards the actual situation, pays attention to the incorrect object, and chooses the wrong option with \texttt{hands} directly. Therefore, it is necessary to eliminate the spurious relation caused by the high or low frequency of co-occurrence through causal learning, and assign the real causal relation between \texttt{baby} and \texttt{strap} to the visual object \texttt{baby}. 
	
	In view of this, Wang et al. propose the VC R-CNN framework to extract the visual commonsense feature \cite{wang2020visual} and strengthen the model's ability of mining correct semantic information by directly concatenating the visual commonsense feature with the original object visual feature. Using the visual object \texttt{baby} in Figure \ref{figure1} as an illustration, the \texttt{baby}'s visual commonsense is made up of different semantic associations with its surroundings, such as \texttt{nipple}, \texttt{adult's hands}, \texttt{bed}, and \texttt{strap}. Once the visual commonsense has been introduced in Figure \ref{figure1} (b), we can guide the model to delve deeper into the semantic information that is relevant to the current situation. We can see that, in comparison to Figure \ref{figure1} (a), the model's confidence in Figure \ref{figure1} (b) for the correct answer "Strap baby to the body" has greatly improved, indicating that the model has taken into account the low-frequency Semantic information \texttt{strap}. Nevertheless, the confidence score is still lower than that of the incorrect answer "Hold her hand". It means that the simple concatenation cannot fully use the inherent potential of the visual commonsense, leading to insufficient interpretability and sub-optimal results.
	
	In this paper, we concentrate on exploring the manner of incorporating the visual commonsense into various scenes, aiming at a plug-and-play module that can be integrated with existing methods. This framework can assist classical VQA models to better transfer and utilize visual commonsense features and improve their reasoning ability. In our method, the visual and textual entities are treated as a jointly constituted heterogeneous graph. The Graph Relation Network (GRN) is introduced to ascertain the correlation between homogeneous edges and the similarity between heterogeneous edges, which updates each entity's relevance dynamically during training. In addition, we generate anchor samples, positive samples, and negative samples based on visual commonsense features and original object visual features for contrastive learning. By bringing the distance between positive and anchor samples closer and the distance between negative and anchor samples farther in the high-dimensional projection space, the model focuses more on visual commonsense features and has better reasoning ability. As shown in Figure \ref{figure1} (c), when leveraging our proposed method, the model focuses on the commonsense knowledge more and retrieves the less frequent semantic information for \texttt{strap} (compared to \texttt{adult's hands}). This phenomenon shows that our method assists the model in concentrating on two key objects (\texttt{baby} and \texttt{strap}), creating a reasonable correlation between them, and making the right prediction. Our module is outfitted with seven exemplary multiple-choice VQA models (\emph{i.e.}, STVQA \cite{jang2017tgif}, CoMem \cite{gao2018motion}, HGA \cite{fan2019heterogeneous}, EIGV \cite{li2022equivariant}, HCRN \cite{le2020hierarchical}, R2C \cite{zellers2019recognition}, and TAB \cite{lin2019tab}) and achieves significant improvements on four datasets (\emph{i.e.}, NExT-QA \cite{xiao2021next}, MSVD-QA \cite{xu2017video}, MSRVTT-QA \cite{xu2017video}, and VCR \cite{zellers2019recognition}).

	\section{Related Work}
	\textbf{Visual Commonsense} has gained a lot of attention in vision-language community \cite{yu2021hybrid,zareian2020learning,wang2020consensus}. Wang et al. \cite{wang2020visual} presented the visual commonsense R-CNN to discover the visual commonsense and concatenate it with the original object feature. Yu et al. \cite{yu2021hybrid} propose a hybrid reasoning network to jointly learn multiple commonsense descriptions for video captioning. Although they take into account the visual commonsense, simple concatenation cannot fully exploit the information contained in visual commonsense. Hence, in this paper, we apply graph contrastive learning to help the model concentrate on visual commonsense and make efficient use of it.\\
	\textbf{Graph-based Models} has been proved effective in tackling the relation reasoning problems recently \cite{liu2020graph,chen2020fine,huang2020location,wang2021dualvgr,gu2021graph}. Approaches based on Graph Convolution Networks \cite{kipf2016semi} show promising results on the Visual Question Answering tasks. Jiang and Han \cite{jiang2020reasoning} builds the model which reasons on the heterogeneous graph and fuses the global and local information to predict the final answer. Xiao et al. \cite{xiao2022video} proposes a conditional graph hierarchy to contextualize the information on different levels. In this paper, we combine Graph Relation Network with contrastive learning, enabling the model to filter out significant objects and learn semantically rich representations.\\
	\textbf{Contrastive Learning} is proposed to learn powerful representations by contrasting positive samples against negative samples, which has been explored in extensive literature \cite{hadsell2006dimensionality,dosovitskiy2014discriminative,DBLP:journals/corr/abs-1808-06670,he2020momentum,zhang2020counterfactual}. Recently, many researchers have attempted to leverage the contrastive learning to solve various vision-and-language tasks and achieved great success. Li et al. \cite{li2020context} introduce a method combining self-attention mechanism with contrastive feature construction, which can summarize common information as well as capture discriminative information. Kim et al. \cite{kim2021self} adpot contrastive learning to contrast the ground truth answer against the other incorrect answers, which assists the model in learning better textual representations. In contrast to these existing approaches, we creatively apply contrastive learning to the visual commonsense in our work, strengthening the model's focus on it.\\
	\textbf{Visual Question Answering} is a crucial area in multimodal machine learning, which has made significant advancements in recent years \cite{yu2018joint,kim2017deepstory,zhao2018multi,li2019learnable,zhang2021explicit,yang2021causal,liu2022show}. The fundamental challenge in these tasks is encoding the input in the visual and linguistic fields and then performing effective multi-modal interaction fusion. To solve the problem, a variety of methods have been proposed during the past few years. Xu et al. \cite{xu2017video} introduces the motion information besides the appearance information in videos to capture the information in the temporal domain. Na et al. \cite{na2017read} and Gao et al. \cite{gao2018motion} adopt memory network in their model to reuse the relevant information. Le et al. \cite{le2020hierarchical} and Dang et al. \cite{dang2021hierarchical} leverage the hierarchical architectures to gain a comprehensive understanding of the videos. On several video question and answer datasets, the aforementioned techniques produced positive outcomes. These techniques still have certain flaws, nevertheless, as a result of the disregard for commonsense knowledge and the efficient use of such information.\\

	\section{Proposed Method}
	\label{headings}
	In this section, we first outline the problem formulation of multiple-choice VQA task. Then we introduce the details of our method, which contains the generation process of contrastive samples, the aggregation of textual and visual features using Graph Relation Network, and the informative feature selection via contrastive learning.

	\subsection{Problem formulation}
	In this part, We take the VQA task of multiple-choice category as an example to describe the spirit of our method. Considering a dataset describing snippets of everyday life, each sample in the dataset contains an image or a video, a question, and several answer candidates. The model is asked to choose the correct one among the answer candidates by incorporating image and text information. Generally, existing methods solve this problem in the following. Firstly, it extracts visual embeddings $F_{V}=\{f^{t_{0}(n)}_{v}\}^{T_{0}(N)}_{t_{0}(n)=1}$, where $T_{0}$ denotes the timestamp in video and $N$ denotes the object number in image, through specific models (\emph{e.g.}, ResNet-152 pretrained on ImageNet \cite{deng2009imagenet}, 3D ResNeXt-101 \cite{hara2018can} pretrained on Kinetics \cite{kay2017kinetics}, or C3D \cite{tran2015learning}). The question and each candidate answer are concatenated respectively and the text encoder (\emph{e.g.}, GloVe \cite{pennington2014glove}, BERT \cite{devlin2018bert}) is then adopted to generate textual embeddings $F_{T}=\{f^{c}_{t}\}^{C}_{m=1}$, where $C$ represents the number of answer candidates. Secondly, the model creates a connection between visual embeddings and textual embeddings by leveraging the multimodal interaction module (\emph{e.g.}, co-attention \cite{yu2019deep}, heterogeneous graph \cite{fan2019heterogeneous}), and outputs the multimodal feature $F_{M}$. Finally, the multimodal feature is fed to the Multilayer Perceptron (MLP) to calculate the final confidence score for each candidate answer. The VQA model is usually optimized by the Cross-Entropy loss to penalize incorrect candidate answers. 
	
\begin{figure*}[t]
		\begin{center}
			\includegraphics[scale=0.0805]{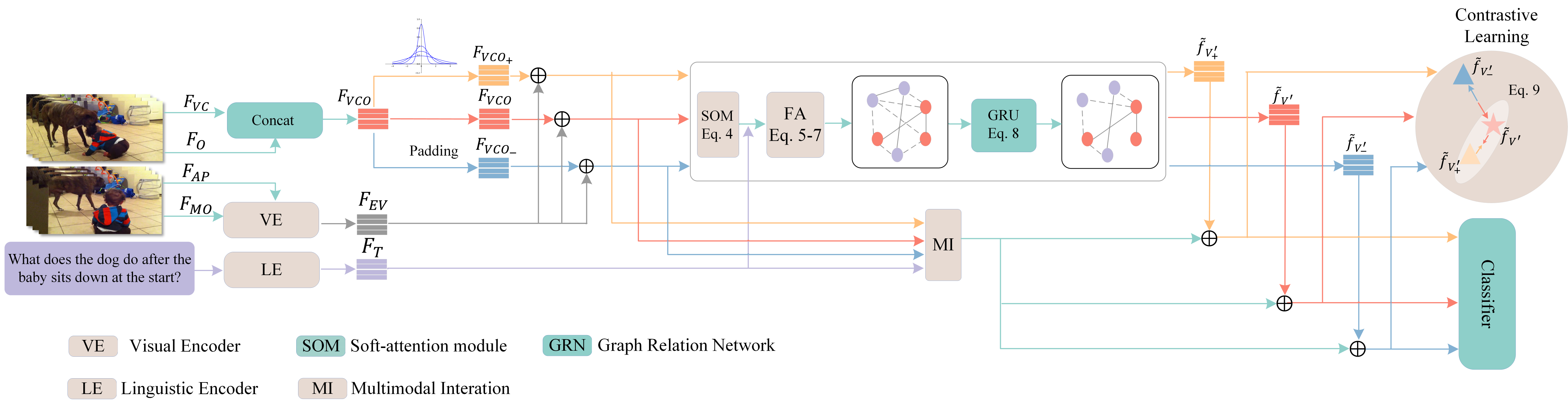}
		\end{center}
		\caption{The overview architecture of our module. The classical VQA pipeline can be separated into three parts: visual encoder (VE), linguistic encoder (LE), and multimodal interaction module (MI). We generate three types of samples: ${F}_{VCO_+}$, ${F}_{VCO}$, ${F}_{VCO_-}$, and incorporate them into the visual encoder. Then, we contextualize the visual embeddings (\emph{i.e.}, ${F}_{VCO}$ and ${F}_{EV}$) and the linguistic embeddings (\emph{i.e.}, ${F}_{T}$) to constitute the heterogeneous graph. The Graph Relation Network (GRN) is introduced to reason about the correlations between various entities of inter-domain and intra-domain. Besides, conditioned on the output of the GRN and the original output of the multimodal interaction module, we design the contrastive learning to guide the model select the discriminative objects. Finally, we employ the Classifier to output the confidence score of all candidate answers.}
		\label{figure2}
	\end{figure*}

	In our approach, we employ a commonsense knowledge extractor (i.e., VC-RCNN \cite{wang2020visual}) to obtain high-level knowledge representation. We extract the visual commonsense feature $F_{VC}=\{f_{vc}^n\}^{N}_{n=1}$ for $N$ objects along with their bounding boxes from each frame, as in \cite{wang2020visual}. By concatenating the commonsense knowledge with the original object feature, we can learn more about the correct relationships rather than the spurious correlations. However, simply concatenating commonsense knowledge is not enough to make the model reason about the right answer. To explicitly develop the model's ability for selecting crucial commonsense knowledge and reasoning about real relationships across various objects, we propose the Visual Commonsense based Heterogeneous Graph Contrastive Learning (VC-HGCL) as a plug-and-play module to equip with a wide range of multimodal methods. The overview architecture of our method is shown in Figure \ref{figure2}.

	\subsection{Contrastive Sample Generating} 
	In this work, we propose a Heterogeneous Graph Contrastive Learning (HGCL) module to utilize commonsense feature $F_{VC}$ with the visual feature $F_{V}$ for the final prediction. It aims to discover the relationship between commonsense knowledge and existing visual features, as well as to equitably distribute the weight of each factor. In this way, our method is supposed to help VQA models select the key object which owns more significant commonsense knowledge for reasoning about the final answer.
	
	The classical VQA pipeline consists of three parts: visual encoder, linguistic encoder, and multimodal interaction module. We first generate the positive, negative, and anchor samples in the visual encoder part. Then, we add our VC-HGCL module to the multimodal interaction module. To be specific, the classical visual encoder module takes images or videos as input and employs feature extractors to obtain the object feature $F_{O}=\{f_{o_{n}}\}^{N}_{n=1}$, the appearance feature $F_{AP}$, and the motion feature $F_{MO}$. In the image-QA tasks, the model generally uses $F_{O}$ and $F_{AP}$; while in the video-QA tasks, since $F_{MO}$ contains temporal information, the model will use $F_{O}$, $F_{AP}$ and $F_{MO}$. The classic video-QA model combines $F_{AP}$ with $F_{MO}$ to obtain enhanced visual feature $F_{EV}$. Following \cite{cho2014learning,le2020hierarchical}, we employ feature processors (e.g. Gated Recurrent Units (GRU) and Conditional Relation Network (CRN)) to generate the enhanced visual feature $F_{EV}$.
	
	Conditioned on the object feature $F_{O}$ and the visual commonsense feature $F_{VC}$, we produce the three types of samples for the consequent contrastive learning. The reason why we employ contrastive learning is as follows: when the visual commonsense feature ($F_{VC}$) and the object visual feature ($F_{O}$) are simply concatenated, $F_{VC}$ receives less weight or even be ignored in some situations. This is due to the fact that $F_{O}$ has a larger dimension than $F_{VC}$ (6144 vs. 1024). In addition, the model will be biased on $F_{O}$, resulting in overfitting. Based on the above facts, we naturally thought of making the model pay more attention to the $F_{VC}$ during the training process, and adaptively balance the weight distribution of $F_{O}$ and $F_{VC}$. Therefore, we introduce the paradigm of contrastive learning. We aggregate the object feature $F_{O}$ and the commonsense knowledge $F_{VC}$ via concatenation to obtain the commonsense-based visual representation ${F}_{VCO}=\{f_{vco_{n}}\}^{N}_{n=1}$, where $N$ represents the number of visual objects in each video frame, which is chosen as the anchor sample:
	\begin{equation}\label{e1}
		\begin{aligned}
			&{F'}_{VCO}=[F_{VC}:F_{O}], \\
			&F_{VCO}= WF'_{VCO}+b, \\
		\end{aligned}
	\end{equation}
	where $W$ is the weight matrix, and $b$ is the bias. To facilitate the calculation of the upcoming contrastive loss, we generate the positive commonsense-based visual representation by disturbing the anchor through the Gaussian Noise. We hope that it will guide the model learn the decisive feature as well as boost the robustness of the model after leveraging contrastive learning. The positive feature can be formulated as:
	\begin{equation}\label{e2}
		\begin{aligned}
			&F_{Gau} \sim E(X)=N(\mu,\sigma^{2}), p(x|\mu,\sigma)=\frac{1}{\sigma \sqrt{2\pi}}e^{-\frac{x}{2\sigma^{2}}}, \\
			&{F'}_{VCO_+}={F'}_{VCO}+F_{Gau}, F_{VCO_+}= W_{+}F'_{VCO_+}+b_{+}, \\ 
		\end{aligned}
	\end{equation}
	where $F_{Gau}$ is sampled from $N(\mu,\sigma^{2})$, $\mu$ is set $0$ and the $\sigma$ is determined by the mean and standard deviation of the anchor feature, and ${F}_{VCO_+}$ denotes the positive commonsense-based visual representation. $W_+$ and $b_+$ represent the parameters involved in the linear transformation.
	
	Contrastively, to impair the informative feature ${F}_{VCO}$ and enforce the model to concentrate more on the commonsense knowledge $F_{VC}$, we discard the $F_{VC}$ from the anchor feature $F_{VCO}$. Specifically, the negative feature can be formulated as:
	\begin{equation}\label{e3}
		\begin{aligned}
			&{F'}_{VCO_-}=F_{O}, \\
			&F_{VCO_-}= W_{-}F'_{VCO_-}+b_{-}, \\
		\end{aligned}
	\end{equation}
	where $W_-$ and $b_-$ represent the parameters involved in the linear transformation. 
	
	The following processing will act on these three types of features (i.e., positive feature ${F}_{VCO_+}$, negative feature ${F}_{VCO_-}$, and anchor feature ${F}_{VCO}$). For convenience, we only use the anchor feature ${F}_{VCO}$ to represent our subsequent processing.
	
	After the aforementioned processing, we add soft-attention module (SOM) for video-QA tasks. The module aims to locate and pick out significant visual elements in each video frame. Once weights have been fairly assigned to various objects, they are aggregated to obtain frame-level feature representations. That is:
	\begin{equation} \label{e4}
		\begin{aligned}
			&F''_{VCO}=\sum^{N}_{i=1}\alpha^{i}_{VCO} f^{i}_{VCO},\\
			&\alpha^{i}_{VCO}= \frac{exp(\sigma(W^{i}_{VCO}f^{i}_{VCO}+b_{VCO}))}{\sum^{N}_{j=1} exp(\sigma(W^{j}_{VCO}f^{j}_{VCO}+b_{VCO}))}, \\
		\end{aligned}
	\end{equation}
	where $W_{VCO}$ is the learnable weight, $b_{VCO}$ is the bias, and $\alpha^{i}_{VCO}$ indicates the importance of each visual object when contextualizing it with the textual information in question and answer candidates. We suppose that the key $f^{i}_{VCO}$ will illuminate the corresponding semantic nodes in the text sentences during the operation of multimodal interaction. Then, we concatenate $F''_{VCO}$ with enhanced visual features $F_{EV}$ and perform feature mapping through a MLP, and the obtained features are recorded as $F''_{EVCO}$. By aggregating the commonsense feature and the enhanced visual feature, we get a more representative image description for detailed visual comprehension.
	
	In order to further aggregate the temporal information of the video frame and the context information of the text, we employ a recurrent neural network to enhance the feature expression. Specifically, we have:
	\begin{equation}\label{e5}
		\begin{aligned}
			&\bar{F}_{EVCO}=RNN (F''_{EVCO}), \\
			&{F'}_{T}=RNN ({F}_{T}),
		\end{aligned}
	\end{equation}
	where we use GRU as the recurrent network. For simplicity, we use $\check{F}_{VCO}$ to uniform the visual feature $\bar{F}_{EVCO}$ in the video-QA tasks and the visual feature $F_VCO$ in the image-QA task.
	
	\subsection{Heterogeneous Graph Contrastive Learning} 
	Now, we proceed to the multimodal interaction module. Conditioned on the visual and textual embeddings, the multimodal interaction module will create connections between the visual and linguistic domains. We insert our proposed module before the final classifier in the multimodal interaction module. We first concentrate on the feature aggregation of $\check{F}_{VCO}$ with textual features ${F'}_{T}$. For all the words in the sentence, we aim to identify the visual features corresponding to the textual information in the image or video through cross-modal interaction, and reinforce the textual features by aggregating the visual features. The calculation process can be expressed as follows:
	\begin{equation}\label{e6}
		\begin{aligned}
			&\beta^{i}_{VCO}= Softmax((W_{QT}F'_{T})(W_{KV}\check{F}_{VCO})^{T}), \\
			&\bar{F'}_{T}=LayerNorm(F'_{T}+FFN_T(\sum^{T_{0}(N)}_{i=1}\beta^{i}_{VCO} W_{VV}\check{F}^{i}_{VCO})), \\
		\end{aligned}
	\end{equation}
	where the superscript $T$ represents the transpose of the matrix, $N$ and $T_{0}$ represent the number of objects in the image and the length of the video sequence, respectively. $W_{QT}$, $W_{KV}$ and $ W_{VV}$ all represent the learnable parameter matrix, $FFN_T$ represents the feed-forward neural network. Similarly, we intend to use visual features in images or videos as queries to locate associated word features in textual information, strengthening visual features by aggregating word features. The calculation process can be expressed as follows:
	\begin{equation}
		\begin{aligned}
			&\beta^{j}_{T}= Softmax((W_{QV}\check{F}_{VCO})(W_{KT}{F'}_{T})^{T}), \\
			&\bar{F}_{VCO}=LayerNorm(\check{F}_{VCO}+FFN_{VCO}(\sum^{M}_{j=1}\beta^{j}_{T}W_{VT}{F'}^{j}_{T})), \\
		\end{aligned}
	\end{equation}
	where the superscript $T$ represents the transpose of the matrix, $M$ represents the number of words in the text sentence. $W_{QT}$, $W_{KV}$ and $W_{VV}$ all represent the learnable parameter matrix.
	
	Conditioned on the aggregated visual feature $\bar{F}_{VCO}$ and textual feature $\bar{F}_{T}$, we view these features as a fully-connected heterogeneous graph $\mathcal{V} = \{\bar{f}^i_{{VCOT}}\}^{T+M}_{i=1} = [\bar{F}_{VCO}:\bar{F}_{T}] = [\{\bar{f}^i_{{VCO}}\}^{T}_{i=1}:\{\bar{f}^j_{{t}}\}^{M}_{j=1}]$ (for video-QA) or $\mathcal{V} = \{\bar{f}^i_{{VCOT}}\}^{N+M}_{i=1} = [\bar{F}_{VCO}:\bar{F}_{T}] = [\{\bar{f}^i_{{VCO}}\}^{N}_{i=1}:\{\bar{f}^j_{{t}}\}^{M}_{j=1}]$ (for image-QA) are associated to the objects/frames and the textual (\emph{i.e.}, question and answer candidates) features, and the edges $\mathcal{E} = \{e^{i,j}\}$ are denoted as the correlations or similarities between node $i$ and $j$. $T$ and $N$ represent the number of frames in video-QA tasks and the number of objects in the image-QA task. Since the correlation and similarity between different nodes in the fully connected heterogeneous graph are inconsistent, each node on the graph also has different contributions to the reasoning of the problem. In view of this, we apply the Graph Relation Network (GRN) to strengthen the cross-modal interaction on the graph, and promote the transfer and utilization of the aggregated visual feature $\bar{F}_{VCO}$ and textual feature $\bar{F}_{T}$. We have:
	\begin{equation}\label{e8}
		\begin{aligned}
			&\mathcal{V} = \{\bar{f}^i_{{VCOT}}\}, i=1,...,N+M, \\
			&e^{i,j}=MLP_R (\bar{f}^{i}_{VCOT},\bar{f}^{j}_{VCOT}), \\
			&IoU^{i,j}=\frac{bbox^i \cap bbox^j}{bbox^i \cup bbox^j}\\
			&\bar{e}^{i,j}=\left\{
			\begin{aligned}
				e^{i,j}, IoU^{i,j} \neq 0\\
				0, IoU^{i,j} = 0\\
			\end{aligned}
			\right.\\
			&e^i=\frac{1}{N_j}\sum_{j\not=i}\bar{e}^{i,j}, \hat{f}^i_{VCOT}=MLP_N (e^i,\frac{1}{M}\sum^M_{i=1}e^i,\bar{f}^i_{VCOT}), \\
			&\hat{\mathcal{V}} = \{\hat{f}^i_{VCOT}\}, i=1,...,T(N), \\
		\end{aligned}
	\end{equation}
	where $MLP_R$ represents non-linear mappings, calculating the values of the edge $e^{i,j}$ adaptively. When the edge is assorted as "heterogeneous edge", the edge information means the semantic similarity that guides the model establish the multimodal alignment. When the edge is classified as "homogeneous edge", the edge represents the relationships between various entities within the same domain, facilitating further inferences. In particular, when applying the GRN in the image-QA task, for two visual objects belonging to the same modality, we further calculate the Intersection over Union (IoU) of their bounding boxes $bbox^i$ and $bbox^j$, which represents whether there is a spatial semantic relationship between two visual objects. When the bounding boxes of two visual objects do not overlap, there is a high probability that there is no interaction between them, so the transmission of information flow between them is cut off. The aggregated feature $e^{i}$ encourages the sharing of knowledge among various entities and domains, which in turn promotes the utilization of commonsense features that are correlated to the surroundings in a variety of contexts. $MLP_R$ also represents non-linear mappings, which is used to update the feature information of the nodes in the heterogeneous graph $\mathcal{G}$.
	
	Then we add residual links that connect $\bar{f}^i_{VCOT}$ and the aggregated node embeddings $\hat{f}^i_{VCOT}$ to obtain the output feature. To assist the following contrastive learning, we employ another soft-attention module to assign weights to reveal the significance of the nodes in the graph. The form of the soft-attention module here is consistent with Equation \ref{e4}, and the output is denoted as $\tilde{F}_{VCOT}$. Furthermore, we concatenate the $\tilde{F}_{VCOT}$ with the multimodal fusion feature $F_{M}$ generated by classical VQA model, and the output is denoted as $F_{out}$. 
	
	Finally, we employ the projector (e.g. MLP) to project the semantic-rich embedding $F_{out}$, and construct a graph contrastive loss similar to InfoNCE \cite{oord2018representation} as follows:
	\begin{equation}\label{e9}
		\mathcal{L}_{cl}=-log\frac{exp(d(\tilde{F}_{V^\prime},\tilde{F}_{V^\prime_+})/\tau)}{exp(d(\tilde{F}_{V^\prime},\tilde{F}_{V^\prime_+})/\tau)+exp(d(\tilde{F}_{V^\prime},\tilde{F}_{V^\prime_-})/\tau)},
	\end{equation}
	where $d(x,y)=x^{T}y/$$\Vert$$x$$\Vert$$\Vert$$y$$\Vert$ represents the dot product between $l_{2}$ normalized $x$ and $y$, $\tilde{F}_{V^\prime}$ represents the projected embedding of the $F_{out}$. $\tilde{F}_{V^\prime}$, $\tilde{F}_{V^\prime_+}$, and $\tilde{F}_{V^\prime_-}$ denotes the anchor, positive, and negative sample that generated through Equation \ref{e6} - \ref{e8} and the second soft-attention module, so that they are originated from $\check{F}_{VCO}$, $\check{F}_{VCO_+}$, and $\check{F}_{VCO_-}$, respectively. We generate $\check{F}_{VCO}$, $\check{F}_{VCO_+}$, and $\check{F}_{VCO_-}$ through Equation \ref{e1} - \ref{e5}, and an aggregation operation of the feature $F''_{VCO}$ and the enhanced visual feature $F_{EV}$. By introducing the loss function Equation \ref{e9}, we induce the model to make the distance in positive pair $(\tilde{F}_{V^\prime},\tilde{F}_{V^\prime_+})$ closer and the distance in negative pair $(\tilde{F}_{V^\prime},\tilde{F}_{V^\prime_-})$ farther. In this way, the model will pay more attention to $F_{VC}$ when performing causal inference and temporal reasoning, and achieve trade-off when assigning weights to $F_{O}$ and $F_{VC}$. Additionally, due to the huge gap in dimension, it is non-trivial to make the distance in positive pair closer than the distance in negative pair.
	
	\begin{table*}[t]
		\caption{\label{table1}The VQA performances of the validation set on NExT-QA before and after applying our module VC-HGCL. The methods with the $\ast$ represent the reproduced results following the official code source. $ACC_{C}$, $ACC_{T}$, and $ACC_{D}$ represent accuracy for causal, temporal, and descriptive questions, respectively.}
		\vspace{-2mm}
		
		\begin{center}
			\begin{tabular}{ c | c c c c | c c c c }
				\hline
				\multicolumn{1}{c| }{\multirow{2}[0]{*}{Methods}}  &  \multicolumn{4}{c| }{\textbf{NExT-QA-GloVe}} &  \multicolumn{4}{c }{\textbf{NExT-QA-BERT}}  \\
				& {\fontsize{8pt}{9.6pt}\selectfont $ACC_{C}$} & {\fontsize{8pt}{9.6pt}\selectfont $ACC_{T}$} & {\fontsize{8pt}{9.6pt}\selectfont $ACC_{D}$} & {\fontsize{8pt}{9.6pt}\selectfont $ACC$} & {\fontsize{8pt}{9.6pt}\selectfont $ACC_{C}$} & {\fontsize{8pt}{9.6pt}\selectfont $ACC_{T}$} & {\fontsize{8pt}{9.6pt}\selectfont $ACC_{D}$} & {\fontsize{8pt}{9.6pt}\selectfont $ACC$} \\
				
				\rowcolor{gray!40}
				STVQA \cite{jang2017tgif} & 36.25 & 36.29 & 55.21 & 39.21 & 44.76 & 49.26 & 55.86 & 47.94 \\
				STVQA$\ast$ & 36.90 & 36.54 & 53.67 & 39.39 & 44.69 & 50.12 & 56.11 & 48.22 \\
				STVQA+Our & $\textbf{37.09}$ & $\textbf{39.08}$ & $\textbf{53.92}$ & $\textbf{40.29}$ & $\textbf{46.11}$ & $\textbf{50.50}$ & $\textbf{56.37}$ & $\textbf{49.12}$ \\
				
				\rowcolor{gray!40}
				CoMem \cite{gao2018motion} & 35.10 & 37.28 & 50.45 & 38.19 & 45.22 & 49.07 & 55.34 & 48.04 \\
				CoMem$\ast$ & 35.44 & 35.30 & 50.06 & 37.67 & 45.80 & 48.33 & 54.57 & 47.98 \\
				CoMem+Our & $\textbf{36.06}$ & $\textbf{36.48}$ & $\textbf{52.90}$ & $\textbf{38.81}$ & $\textbf{47.10}$ & $\textbf{49.75}$ & $\textbf{56.50}$ & $\textbf{49.42}$ \\ 
				
				\rowcolor{gray!40}
				HGA \cite{fan2019heterogeneous} & 35.71 & 38.40 & 55.60 & 39.67 & 46.26 & 50.74 & 59.33 & 49.74 \\
				HGA$\ast$ & 36.98 & 39.08 & 55.73 & 40.57 & 46.68 & 49.81 & 57.53 & 49.38 \\
				HGA+Our & $\textbf{38.93}$ & $\textbf{41.32}$ & $\textbf{56.24}$ & $\textbf{42.39}$ & $\textbf{47.91}$ & $\textbf{50.81}$ & $\textbf{58.69}$ & $\textbf{50.52}$ \\
				
				\rowcolor{gray!40}
				EIGV \cite{li2022equivariant} & - & - & - & - & - & - & - & - \\
				EIGV$\ast$ & 42.31 & 43.49 & 62.03 & 45.76 & 49.83 & 52.92 & 63.06 & 52.88 \\
				EIGV+Our & $\textbf{44.61}$ & $\textbf{45.35}$ & $\textbf{62.42}$ & $\textbf{47.62}$ & $\textbf{51.05}$ & $\textbf{53.29}$ & $\textbf{63.45}$ & $\textbf{53.70}$ \\ 
			\end{tabular}
		\end{center}
	\end{table*}
	
	\begin{table*}[t]
		\caption{\label{table2}The VQA performances of the testing set on NExT-QA before and after applying our module VC-HGCL. The methods with the $\ast$ represent the reproduced results following the official code source. $ACC_{C}$, $ACC_{T}$, and $ACC_{D}$ represent accuracy for causal, temporal, and descriptive questions, respectively.}
		\vspace{-2mm}
		
		\begin{center}
			\begin{tabular}{ c | c c c c | c c c c }
				\hline
				\multicolumn{1}{c| }{\multirow{2}[0]{*}{Methods}}  &  \multicolumn{4}{c| }{\textbf{NExT-QA-GloVe}} &  \multicolumn{4}{c }{\textbf{NExT-QA-BERT}}  \\
				& {\fontsize{8pt}{9.6pt}\selectfont $ACC_{C}$} & {\fontsize{8pt}{9.6pt}\selectfont $ACC_{T}$} & {\fontsize{8pt}{9.6pt}\selectfont $ACC_{D}$} & {\fontsize{8pt}{9.6pt}\selectfont $ACC$} & {\fontsize{8pt}{9.6pt}\selectfont $ACC_{C}$} & {\fontsize{8pt}{9.6pt}\selectfont $ACC_{T}$} & {\fontsize{8pt}{9.6pt}\selectfont $ACC_{D}$} & {\fontsize{8pt}{9.6pt}\selectfont $ACC$} \\
				
				\rowcolor{gray!40}
				STVQA \cite{jang2017tgif} & - & - & - & - & 45.51 & 47.57 & 54.59 & 47.64 \\
				STVQA$\ast$ & 34.87 & 35.79 & 54.52 & 38.38 & 45.87 & 48.40 & 54.52 & 48.07 \\
				STVQA+Our & $\textbf{36.58}$ & $\textbf{37.30}$ & $\textbf{54.59}$ & $\textbf{39.75}$ & $\textbf{47.53}$ & $\textbf{49.23}$ & $\textbf{56.30}$ & $\textbf{49.50}$ \\ 
				
				\rowcolor{gray!40}
				CoMem \cite{gao2018motion} & - & - & - & - & 45.85 & 50.02 & 54.38 & 48.54 \\
				CoMem$\ast$ & 35.63 & 34.32 & 51.60 & 37.84 & 44.60 & 47.38 & 54.66 & 47.12 \\
				CoMem+Our & $\textbf{36.38}$ & $\textbf{38.05}$ & $\textbf{54.59}$ & $\textbf{39.89}$ & $\textbf{47.16}$ & $\textbf{49.57}$ & $\textbf{55.87}$ & $\textbf{49.33}$ \\ 
				
				\rowcolor{gray!40}
				HGA \cite{fan2019heterogeneous} & - & - & - & - & 48.13 & 49.08 & 57.79 & 50.01 \\
				HGA$\ast$ & 38.56 & 38.95 & 56.65 & 41.56 & 47.47 & 48.63 & 58.01 & 49.56 \\
				HGA+Our & $\textbf{39.03}$ & $\textbf{40.61}$ & $\textbf{58.08}$ & $\textbf{42.64}$ & $\textbf{48.38}$ & $\textbf{50.66}$ & $\textbf{58.29}$ & $\textbf{50.71}$ \\ 
				
				\rowcolor{gray!40}
				EIGV \cite{li2022equivariant} & - & - & - & - & - & - & - & 53.7 \\
				EIGV$\ast$ & 42.58 & 42.00 & 60.57 & 45.35 & 50.20 & 51.94 & 61.99 & 52.67 \\
				EIGV+Our & $\textbf{43.89}$ & $\textbf{45.24}$ & $\textbf{61.49}$ & $\textbf{47.20}$ & $\textbf{50.89}$ & $\textbf{52.20}$ & $\textbf{62.21}$ & $\textbf{53.15}$ \\ 
				
			\end{tabular}
		\end{center}
	\end{table*}

	\subsection{Answer Prediction}
	In this paper, we focus on multiple-choice tasks, in which the model is supposed to choose the correct answer from several candidate answers $\{A_{i}\}^M_{i=1}$. We follow the traditional processing flow \cite{jang2019video} to combine questions with each candidate. We adopt linear regression function as Classifier that takes the output feature $F_{out}$ as input and outputs scores for each candidate answer. We employ a hinge-based loss on the prediction scores of the answer candidates as follows:
	\begin{equation}\label{e10}
		\mathcal{L}_{pre}=\sum_{i=1}^{M-1}max(0,1+y^{n}_{i}-y^{p}),
	\end{equation}
	where $y^{n}_{i}$ and $y^{p}$ represents incorrect answer candidates and correct answer candidate, respectively. Finally, taking Equation \ref{e9} and Equation \ref{e10} into account, we train the whole network with the following loss function:
	\begin{equation}
		\mathcal{L}_{whole}=\mathcal{L}_{pre}+ \lambda\mathcal{L}_{cl},
	\end{equation}
	where $\lambda$ is the trade-off parameter, which is used to adjust the graph contrastive loss to the same order of magnitude as the hinge-based loss. 
	
	\section{Experiments}
	\subsection{Datasets} 
	We evaluate our method on four multiple-choice VQA benchmarks including three video-QA datasets and one image-QA dataset, which demonstrates the effectiveness and generalizability of our method thoroughly. We follow the training and evaluation protocol in the original paper in all four benchmarks.
	
	\textbf{NExT-QA} \cite{xiao2021next} is a video-QA dataset that goes beyond descriptive QA to causal and temporal action reasoning. Besides, the videos in NExT-QA all reflect real daily life, consisting of a wealth of information. The dataset contains 5,440 videos and 52,044 Question-Answer(QA) pairs, which are split into 34,132, 4,996, and 8,564 QA pairs for the training, validation, and testing set, respectively. 
	
	\textbf{MSVD-QA} and \textbf{MSRVTT-QA} \cite{xu2017video} are two video-QA datasets targeting the recognition problems, which includes five different types of questions: $what, who, how, when$, and $where$. The two datasets are generated through an automatic method. MSVD-QA dataset has 1,970 video clips and 50,505 QA pairs. The database is split to 61\%, 13\%, and 26\% for the training, validation, and testing set, respectively. MSRVTT-QA dataset describes more complex scenes and contains 10K videos and 243k QA pairs. The training, validation, and testing set take 65\%, 5\%, and 30\% of the total number of videos.
	
	\textbf{VCR} \cite{zellers2019recognition} is a representative large-scale image-QA dataset which explores the visual commonsense reasoning. The task in the dataset can be decomposed into two sub-tasks: (1) Question answering (Q $\rightarrow$ A) supposes the model to predict the answer conditioned on the question; (2) Answer justification (QA $\rightarrow$ R) needs the model to choose the rationale answer to justify the predicted answer in Q $\rightarrow$ A subtask. The dataset contains 290K multiple choice problems derived from movie scenes. We use 80,418 images with 212,923 questions for training, and 9,929 images with 26,534 questions for testing.
	
	\subsection{Implementation Details}
	We integrate our module into seven representative multiple-choice VQA models to demonstrate the effectiveness of our module. It consists of STVQA \cite{jang2017tgif}, CoMem \cite{gao2018motion}, HCRN \cite{le2020hierarchical}, HGA \cite{fan2019heterogeneous}, EIGV \cite{li2022equivariant}, HCRN \cite{le2020hierarchical}, R2C \cite{zellers2019recognition}, and TAB \cite{lin2019tab} with available source codes. The rest is identical to the original approach, except for adding our modules where necessary. As for hyperparameters, such as learning rate, batch size, etc., we all follow the configuration of the source code.
	
	\subsection{Quantitative Results}
	\subsubsection{Results on NExT-QA}
	Tables~\ref{table1} and~\ref{table2} exhibit the detailed results of the validation set and testing set on NExT-QA datasets. For each VQA model, we list three results: the results reported in the original paper, the reproduced results following the official code source, and the results incorporated with our VC-HGCL module. 
	
	First, we can see that the reproduced results achieve similar results to the original paper. This ensures that our results are fair and credible. 
	
	Second, from both tables, we observe that VC-HGCL improves the performance in all representative VQA models of all the subtasks on NExT-QA, indicating that our method is agnostic to methods and can be a plug-and-play module to equip with all the popular VQA models. On the whole, when using the glove feature, the boost of the model performance after incorporating our VC-HGCL module ranged from $0.90\%$ to $1.86\%$ and $1.08\%$ to $2.05\%$ in the aspect of validation set and testing set under the average metric. When using the BERT feature, the performance gap between the enhanced model and the baseline model reaches $0.82\%$ to $1.44\%$ and $0.48\%$ to $2.21\%$. The results imply that by utilizing commonsense knowledge to identify important objects and retrieve semantic information relevant to the current scene, our module fosters the interaction between the visual and linguistic domains. 
	
	Third, by carefully examining the $Acc_{C}$ and $Acc_{T}$ columns of these two tables, we find that our module significantly improves the performance of the Causal subtask and the Temporal subtask. To be specific, when using the glove feature, the performance gains of up to $2.30\%$ and $3.73\%$ improvements in the Causal and Temporal subtask, respectively. When using the BERT feature, the two settings witness performance increases of up to $2.56\%$ and $2.19\%$, respectively. The above performance improvements demonstrate that our method guides the model pay more attention to the relevant objects that provide key commonsense information. Our method not only establishes efficient and reliable information flow between commonsense information, object visual features, and textual features, but also promotes causal reasoning ability in heterogeneous graphs. We want to emphasize that the number of question-answer pairs in the Causal subtask is twice and four times that of the Temporal subtask and Descriptive subtask, respectively. This further proves that our module can reasonably and effectively capture the underlying commonsense information and perform cognitive-level reasoning, rather than just staying at the perceptual level.
	
	Furthermore, we observe that when our module is used on a more complex and robust baseline model (\emph{e.g.}, CoMem and HGA), it achieves a significant performance improvement compared to the simple baseline model. This is due to the fact that when the model has a strong ability to interact with visual and textual information, the model is superior at inducing Graph Contrastive Learning to find the key visual commonsense features in the projected space. These visual commonsense features can be mined in the visual-text information flow of heterogeneous graphs, and further feedback to strengthen the weight of visual and textual entities that play an important role in reasoning and cognition.

	\begin{figure}
		\begin{center}
			\includegraphics[scale=0.135]{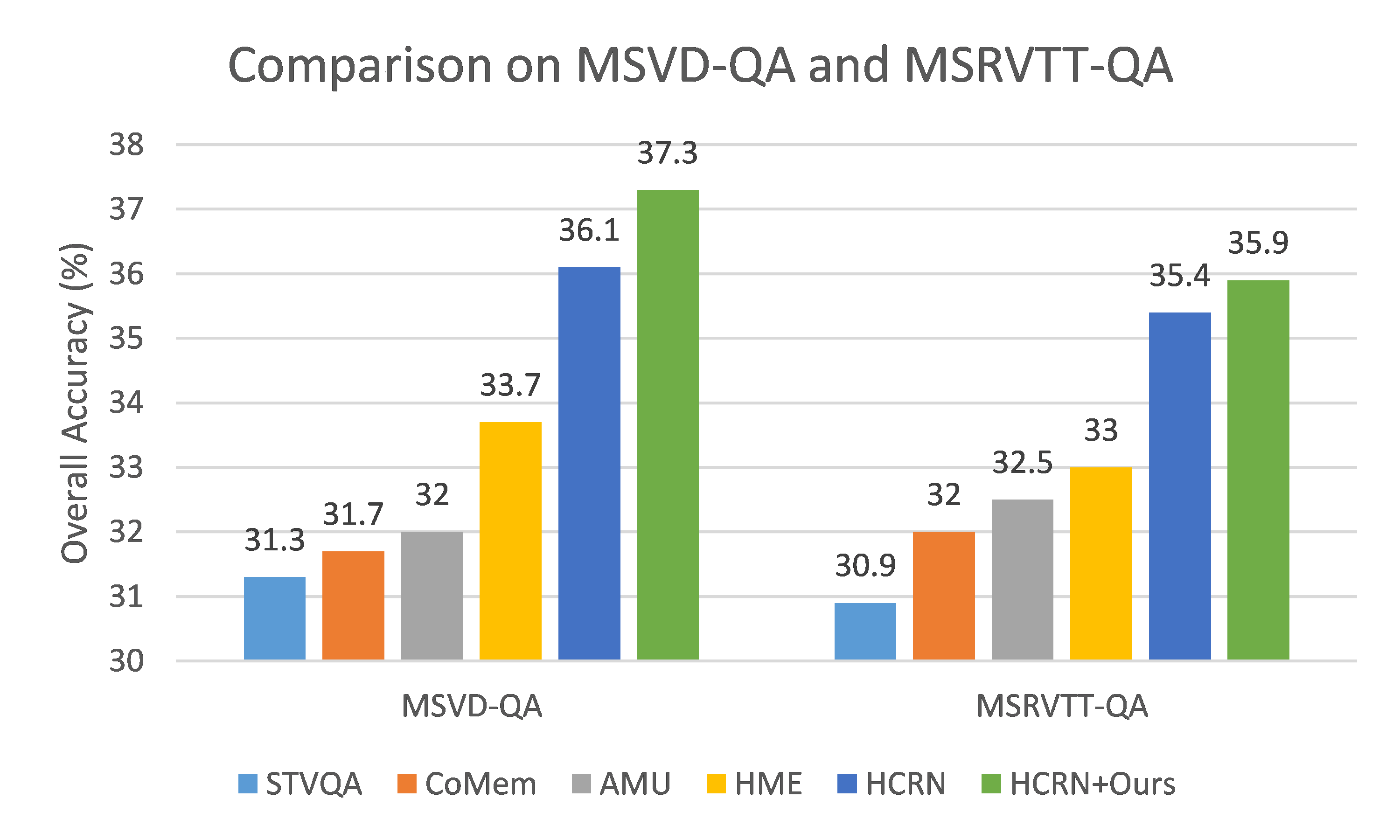}
		\end{center}
		\caption{Comparison with other methods on MSVD-QA and MSRVTT-QA dataset. The value is the accuracy (\%).}
		\label{figure3}
	\end{figure}
	
	\subsubsection{Results on MSRVTT-QA and MSVD-QA}
	Figure \ref{figure3} reports the results in MSRVTT-QA and MSVD-QA datasets. It can be included from Figure \ref{figure3} that our proposed model helps the HCRN method achieve $37.3\%$ and $35.9\%$ accuracy, outperforming the baseline model by 1.2 points and 0.5 points on MSVD-QA and MSRVTT-QA dataset, respectively. The results demonstrate that adding a graph contrastive learning architecture with commonsense knowledge assists the model in improving the association between important visual and textual nodes in the multimodal interaction graph. Additionally, the fact that the scale of MSRVTT-QA is roughly five times larger than that of MSVD-QA illustrates how well our module handles both large and small scale datasets.
	
	\subsubsection{Results on VCR}
	Table 3 displays three different sorts of outcomes on two classic models: the initial results, the results we replicate, and the results after incorporating our module. As implied by its name, VCR (Visual Commonsense Reasoning) is a dataset that places great demands on a model's ability to reason about the visual commonsense. We have seen encouraging improvement of our method on this dataset. To be specific, on the $Q \rightarrow A$ and $QA \rightarrow R$ subtasks, the model surpasses the baseline model by $1.2\%$ to $5.2\%$ and $0.8\%$ to $2.8\%$, respectively. It is also noteworthy that the model outperforms the most challenging task $Q \rightarrow AR$ by $1.5\%$ to $5.1\%$. This indicates that our module effectively assists VQA models in learning the most important knowledge information in the complex visual features, regardless of how basic or sophisticated the model is. In order to get a more thorough cognitive-level understanding of the reasoning scene, engagement with textual information conditioning on the important visual information is further encouraged by our method.
	
	\begin{table}[H]
		\centering
		\caption{\label{table3}The accuracy for three subtasks on VCR before and after applying VC-HGCL.}
		\vspace{-2mm}
		\begin{tabular}{p{1.7cm}<{\centering} | p{1.60cm}<{\centering} p{1.60cm}<{\centering} p{1.60cm}<{\centering} }
			\hline
			Method & $Q \rightarrow A$ & $QA \rightarrow R$ & $Q \rightarrow AR$ \\
			\hline
			\rowcolor{gray!40}
			R2C \cite{zellers2019recognition} & 63.8 & 67.2 & 43.1  \\
			R2C$\ast$ & 63.6 & 67.9 & 43.7 \\
			R2c+Ours & $\textbf{68.8}$ & $\textbf{70.7}$ & $\textbf{48.8}$ \\
			\rowcolor{gray!40}
			TAB \cite{lin2019tab} & 69.5 & 71.6 & 50.1  \\
			TAB$\ast$ & 69.3 & 71.2 & 49.5  \\
			TAB+Ours & $\textbf{70.5}$ & $\textbf{72.0}$ & $\textbf{51.0}$ \\
			\hline
		\end{tabular}
	\end{table}
	
	\subsection{Ablation Study}
	To investigate how each submodule contributes to the final results, we design ablation studies about four variants and evaluate them via the strong baseline model HGA on the NExT-QA dataset of the validation set. We use the GloVe for textual embeddings. First, as shown in Table \ref{table4}, concatenating the object feature and the visual commonsense feature slightly improves model performance from 40.57 to 40.77. However, concatenation is too simple to make the model treat both features. Therefore, this is not a sensible or effective use of the visual commonsense information. Second, we leverage the Contrastive Learning to select the discriminative features and learn effective representations for the complex visual and textual content. Results of the second row and the third row point out the crucial of the Contrastive Learning. It demonstrates that, even though we merely employ a straightforward MLP structure, the model is capable of superficially selecting vital commonsense features and giving them more weight. Finally, beyond MLP, we present the Graph Relation Network (GRN) to help the Contrastive Learning be more effective. The Graph Contrastive Learning module enables intra-modal and inter-modal information transfer and reasoning in the visual domain and linguistic domains. As a result, the model not only has a thorough comprehension of the scene but also acquires guiding information from related objects. For reasoning in cross-modal interactions, it also collects implicit commonsense information from the text side. By comparing the results in the last two rows of Table \ref{table4}, our above conjecture is well verified.
	
	\begin{figure}
		\begin{center}
			\includegraphics[scale=0.07]{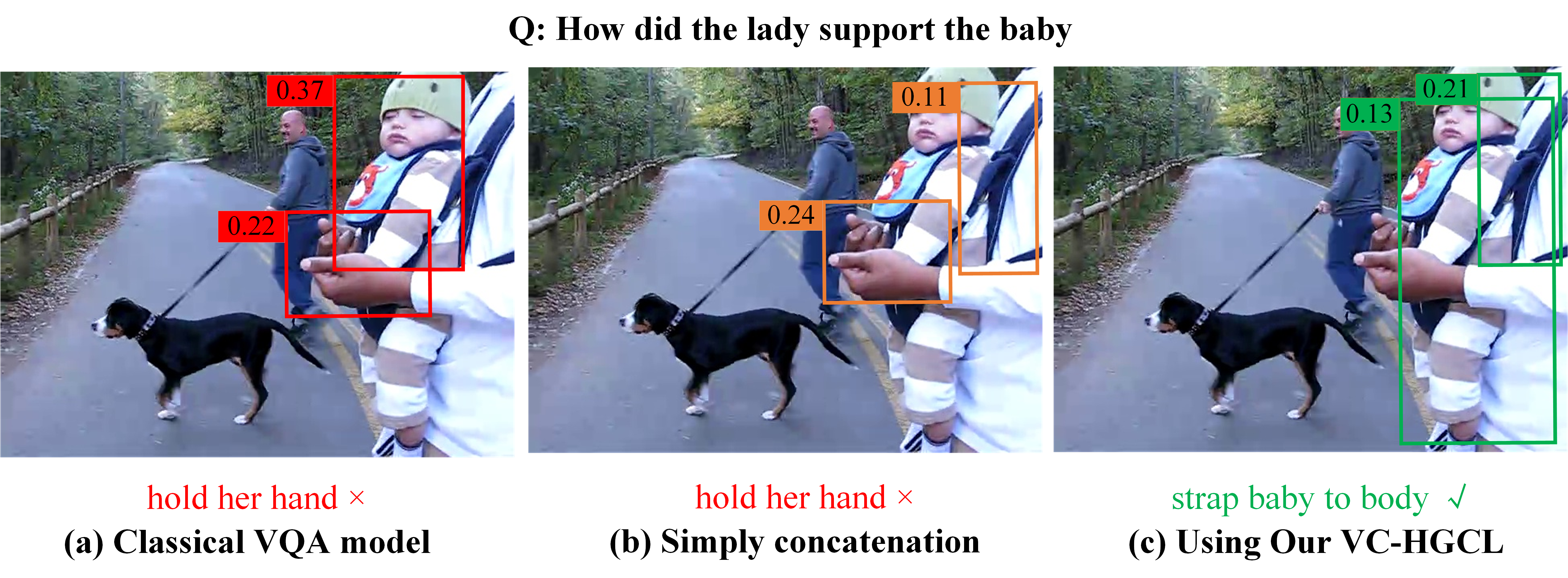}
		\end{center}
		\caption{The attention areas and the corresponding weights predicted by the classical VQA model, the VQA model that simply concatenated with visual commonsense, and our VC-HGCL module, respectively.}
		\label{figure4}
	\end{figure}
	
	\subsection{Qualitative Analysis}
	In order to provide a thorough study of how our module fairly allocates weights to the pertinent important items, we show the attention areas for SOM (Eq. \ref{e4}). An example from the NExT-QA dataset illustrates the varying focus areas of several approaches in Figure \ref{figure4}. The red, orange, and green boxes represent the attention areas predicted by the classical VQA model, the VQA model that simple concatenated with visual commonsense, and our module, respectively. We notice that the classical VQA model mostly concentrates on regions such as \texttt{baby} and \texttt{adult's hand} since "Hold the baby" is a common scene in daily life. After introducing visual commonsense, the model takes into account the correlations with the surroundings in diverse contexts. Once our module has been incorporated, the model focuses even more on low-frequency visual components that are appropriate for this scene and gives them greater weights, like \texttt{adult's body} and \texttt{strap}. This outcome demonstrates how our module can attend to more reasonable areas and enhance the model's performance.
	
	\begin{table}[t]
		\begin{center}
			\caption{\label{table4}Ablative results of the proposed method on validation set of NExT-QA . The experiments are conducted on HGA using GloVe embeddings. $ACC_{C}$, $ACC_{T}$, and $ACC_{D}$ represent accuracy for causal, temporal, and descriptive questions, respectively. VCO, MLP, GRN represents concatenation of $F_{VCO}$, Multilayer Perceptron with contrastive learning, and Graph Relation Network with contrastive learning, respectively.}
			\vspace{-2mm}
			\begin{tabular}{lccccccc}
				\toprule
				\multicolumn{3}{c}{Component}  & \multirow{2}{0.8cm}{$ACC_{C}$} & \multirow{2}{0.8cm}{$ACC_{T}$} & \multirow{2}{0.8cm}{$ACC_{D}$} & \multirow{2}{0.8cm}{$ACC$}\\
				\cmidrule(lr){1-3}  VCO & MLP & GRN \\
				\toprule
				&   &  & 36.98 & 39.08 & 55.73 & 40.57 \\
				$\checkmark$   & & & 37.36  & 40.20 & 53.41 & 40.77    \\
				$\checkmark$   &  $\checkmark$ &  &  37.78 &  41.07 & 55.98 & 41.67  \\
				$\checkmark$   &  & $\checkmark$ & \textbf{38.93}  & \textbf{41.32} & \textbf{56.24} & \textbf{42.39} \\
				\bottomrule
			\end{tabular}
		\end{center}
	\end{table}
	
	\section{Conclusion}
	In this paper, we delve into visual question answering and reveal the weaknesses of existing approaches from the visual commonsense perspective. We suggest the Visual Commonsense based Heterogeneous Graph Contrastive Learning (VC-HGCL) module to more effectively employ the visual commonsense features to improve the model's capacity for prediction and interpretability. Additionally, our module can be integrated with various VQA models as a model-agnostic plug-and-play module. We transfer the visual commonsense features and textual features by building a dynamic entity correlation graph. We use it to dynamically correct the selection of discriminative objects in the scene and the derivation of entity correlation or similarity. Our extensive experiments over various benchmarks on seven representative VQA methods indicate the effectiveness and generalizability of the proposed method. In future work, we will generalize our module to more vision-language tasks, such as image captioning and visual dialog.
	
	\bibliographystyle{ACM-Reference-Format}


\end{document}